\documentclass[conference]{IEEEtran}
\usepackage{cite}
\usepackage{amsmath,amssymb,amsfonts}
\usepackage{algorithm}
\usepackage{algorithmic}
\usepackage{graphicx}
\usepackage{textcomp}
\usepackage{xcolor}
\usepackage{multirow}
\usepackage{pgfplots}

\def\BibTeX{{\rm B\kern-.05em{\sc i\kern-.025em b}\kern-.08em
    T\kern-.1667em\lower.7ex\hbox{E}\kern-.125emX}}
\begin{document}

\title{
Class-incremental Learning using a Sequence of Partial Implicitly Regularized Classifiers
}

\author{

\IEEEauthorblockN{1\textsuperscript{st} Sobirdzhon Bobiev}
\IEEEauthorblockA{\textit{Institute of Data Science and Artificial Intelligence} \\
\textit{Innopolis University}\\
Innopolis, Russia, 420500 \\
s.bobiev@innopolis.university}
\and
\IEEEauthorblockN{2\textsuperscript{nd} Adil Khan}
\IEEEauthorblockA{\textit{Institute of Data Science and Artificial Intelligence} \\
\textit{Innopolis University}\\
Innopolis, Russia, 420500 \\
a.khan@innopolis.ru}
\and
\IEEEauthorblockN{3\textsuperscript{rd} Syed Muhammad Ahsan Raza Kazmi}
\IEEEauthorblockA{\textit{Institute of Secure and Cyber Physical System} \\
\textit{Innopolis University}\\
Innopolis, Russia, 420500 \\
a.kazmi@innopolis.ru}
}

\maketitle

\begin{abstract}
In class-incremental learning, the objective is to learn a number of classes sequentially without having access to the whole training data. However, due to a problem known as \textit{catastrophic forgetting}, neural networks suffer substantial performance drop in such settings. The problem is often approached by \textit{experience replay}, a method which stores a limited number of samples to be replayed in future steps to reduce forgetting of the learned classes. When using a pretrained network as a feature extractor, we show that instead of training a single classifier incrementally, it is better to train a number of specialized classifiers which do not interfere with each other yet can cooperatively predict a single class. Our experiments on CIFAR100 dataset show that the proposed method improves the performance over SOTA by a large margin.
\end{abstract}


\begin{figure*}
    \centering
    \includegraphics[]{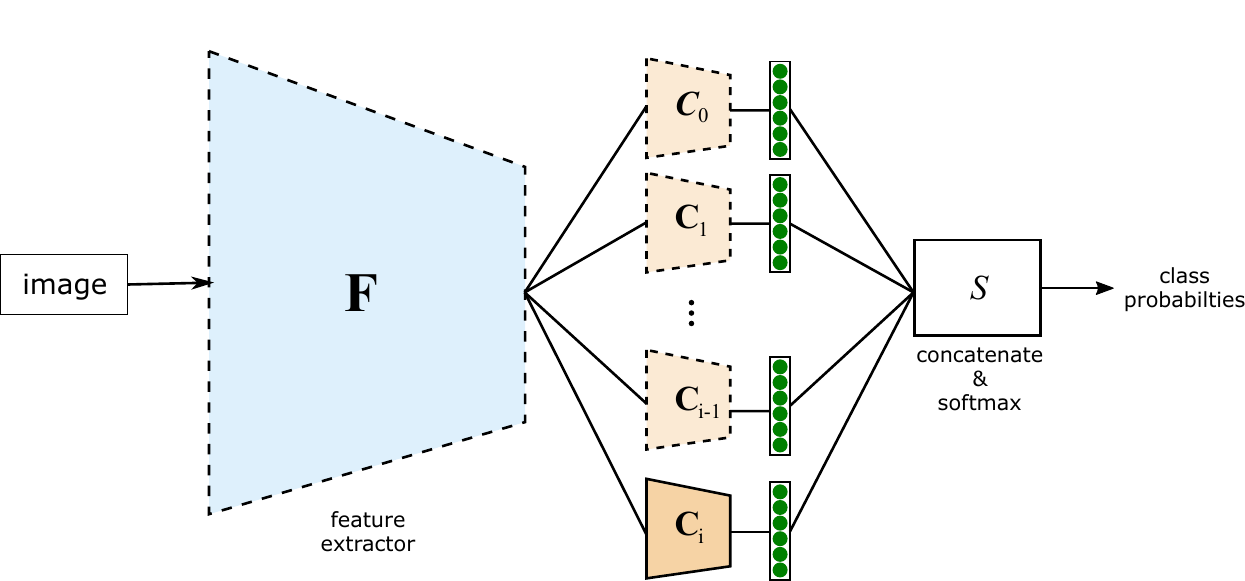}
    \caption{Model architecture. It consists of a feature extractor $\textbf{F}$ (pale-blue trapezoid) and a growing number of classifiers (orange trapezoids), followed by the final layer $S$ which concatenates all classifier outputs together and applies the softmax function. At each training session $i$, only the newly added classifier $\textbf{C}_i$ is trained while the previous classifiers and the feature extractor are kept frozen. The frozen modules are represented by dashed borders.}
    \label{fig:diagram}
\end{figure*}

\section{Introduction}
Artificial neural networks(ANNs) have been at the top of the machine learning landscape for a while. They have achieved impressive performances across various applications, including object recognition\cite{khan2020post}, anomaly detection\cite{rivera2020anomaly, yakovlev2020abstraction}, accident detection\cite{batanina2019domain, bortnikov2019accident}, action recognition\cite{gavrilin2019across, sozykin2018multi, khan2010accelerometer}, scene classification\cite{protasov2018using}, hyperspectral image classification\cite{ahmad2020fast, ahmad2019multi}, medical image analysis\cite{gusarev2017deep, dobrenkii2017large}, machine translation\cite{khusainova2019sart, valeev2019application} etc. Their success is mainly attributed to the availability of large amounts of data and sufficient computational power. While ANNs are inspired by the biological brain, still there are some shortcomings that makes them different. Specifically, they are not designed to learn in incremental way, like humans do. Humans keep learning new knowledge throughout their lives. However, experiments show that ANNs almost completely forget their previous knowledge when they are trained on a new task\cite{kemker2017review}. This is termed as ``Catastrophic Forgetting"\cite{mccloskey1989}. 

The common way to train ANNs is to provide them the whole dataset at once, and let them iterate through it multiple times. Unfortunately, there are situations where this is not a feasible option. There can be memory limits in the learning device, making it impossible to store all data. Also, there can be security concerns for storing the data if it contains sensitive information. Catastrophic forgetting in ANNs was first addressed by McCloskey\cite{mccloskey1989} in 1989, but is still an unsolved problem hindering the progress towards building AI agents that can learn continuously. 

Three main continual learning scenarios are identified: task-incremental learning, class-incremental learning, and domain-incremental learning \cite{vandeven2019scenarios}. In task-incremental learning, the model is required to learn a sequence of tasks sequentially, and during inference it will be provided with the task identity. On the other hand, in class-incremental learning, no such information is provided at inference time, and the model is required to predict both the correct class and the task. A slightly different scenario is domain-incremental learning, which, unlike class-incremental learning, does not require predicting the task identity.

This paper is concerned about class-incremental learning, where at each training session multiple new classes are to be learned while also maintaining the knowledge previous classes. Despite many proposed solutions in the literature, the baseline methods like ``Experience Replay" (ER)\cite{buzzega2020rethinkingER} and GDumb\cite{prabhu2020greedy} are still shown to be as effective as the state of the art. Given that each training session contains examples of only a few classes, the deep networks are prone to overfitting. Therefore, it is plausible to use a frozen feature extractor and smaller classifiers on top. ER trains only a single network, extending its outputs units for the new classes. In each training session, the network is trained on the examples of new classes as well as a small number of examples from previous classes which have been stored in the limited memory buffer. While retraining on these few stored examples helps retain the knowledge of past classes, there is still no guarantee that the new knowledge will not interfere with past. To solve this problem, we propose to train a separate classifier for each group of new classes. These classifiers do not share any weights with each other, implying that the newly acquired knowledge will be stored separately without overriding others. 

\section{Related Work}

In the following, we present some of the continual learning method existing in the literature. We divide these methods into three groups which mainly rely on one of the three ideas: replay, regularization, and architectural techniques, which are all described in the following subsections.

\subsection{Replay Methods}
A straightforward solution to prevent catastrophic forgetting is to revisit the previous tasks. Rehearsal methods accomplish this by storing a small portion of the seen examples for later retraining. When the model is faced with new training data, it augments it with memory samples to reduce the forgetting of previous knowledge. 

iCaRL\cite{Rebuffi2017iCaRL} uses the  nearest-mean-of-exemplars classification approach, classifying items to the class with the nearest center. 
It uses a heuristic approach for updating the memory buffer, prioritizing items based on their proximity to their class mean. Together with network distillation, it has been able to learn in class-incremental learning scenario. Gradient Episodic Memory (GEM)\cite{lopez2017GEM} is designed for task-incremental learning from streaming data. In this setting, the model receives a series of tuples $(x, t, y)$ consisting of input, task label, and target, respectively. GEM uses the memory samples not for replay but to serve the inequality constraints that prevents the loss for the past tasks to decrease. Replay using Memory Indexing (REMIND) \cite{hayes2020remind} applies quantization to the extracted feature maps from deeper layers of a CNN and stores their indices in memory. This results in a compressed representation and allows for much more number of stored examples. REMIND's architecture consists of a frozen feature extractor followed by a trainable classifier. REMIND was shown to outperform existing approaches in a streaming setup on the datasets ImageNet, and CORe50\cite{lomonaco2017core50}.

\paragraph{Rethinking Experience Replay\cite{buzzega2020rethinkingER}} 
The authors emphasize that \textit{simple experience replay}(or simple rehearsal)\cite{ratcliff1990} is still as effective as state of the art if certain issues are resolved. As stated by the authors, the three important issues with this approach are: overfitting to stored examples, biased prediction and accuracy towards the later classes, and non-i.i.d stored data in cases where the buffer is small. Their proposals include an additional bias correction layer (BiC)\cite{wu2019BiC}, exponential decay of learning rate, and balanced sampling for the memory buffer. In their reported results ER has outperformed SOTA sophisticated replay methods such as iCaRL\cite{Rebuffi2017iCaRL}, GEM\cite{lopez2017GEM}, A-GEM\cite{chaudhry2018A-GEM}, and HAL\cite{chaudhry2020HAL}, sometimes by a very large margin. 

\paragraph{GDumb \cite{prabhu2020greedy}} is a recent paper that questioned our progress in continual learning. It proposes a simple and most generic baseline for continual learning. It basically maintains a balanced memory buffer and only trains on the samples contained in it. The reported results have shown that it outperforms many SOTA methods in their respective settings for which they were designed to. GDumb serves a strong baseline for all continual settings including class-incremental learning.

\subsection{Regularization Methods}
These methods impose a type of regularization that helps retain the learned knowledge of past tasks. They do it by adding additional loss terms to the loss function.

Learning without forgetting (LwF)\cite{li2017LWF} presented a modification to the standard fine-tuning. A new network is initialized as a copy of the old one with an extension of the output layer(called \textit{multihead} approach) for the new task. A distillation loss is added between old and new task heads so the new network is reminded of the old tasks indirectly. In backward pass, only the new network is updated. 

Elastic Weight Consolidation (EWC)\cite{kirkpatrick2017EWC} and Synaptic Intelligence (SI) \cite{zenke2017SI} both try to approximate the importance of each of the parameters for previous tasks and selectively apply regularization to limit their change. Regularization methods alone, as several papers demonstrate, are not sufficient for proper class-incremental learning\cite{kemker2017review, lesort2020regularization}.

\subsection{Architectural Methods}
Architectural methods try to manipulate weights, neurons, layers, or architecture of the network to protect the learned knowledge while acquiring a new one. They either use fixed or dynamic architectures. These methods have the advantage of completely eliminating interference between tasks while allowing knowledge transfer between them. However, practically, they are coupled with scalability issues as the number of tasks grow. 

``Progressive Neural Networks"(PNN)\cite{rusu2016PNN} inherently target the task-incremental learning. Their architecture grows laterally, each time adding a new neural network, called ``columns", for a new task. The new columns have connections to other layers of previous networks and thus, highly benefit from knowledge transfer. Once a column is trained, it will be kept frozen making the PNNs completely immune to catastrophic forgetting. A big issue with PNNs is that they keep growing too large, limiting their use in practice to only a small number of tasks. ``Compacting, Picking and Growing" (CPG) \cite{hung2019CPG} tries to overcome this issue by dynamically controlling the architecture of the network in an efficient way. The weights of the network are grouped into ``compact" and ``free", where the newly added weights are considered free until they are trained and then compacted. For each new task, a learnable binary mask is created and applied to the compact weights to select a subset of them. Then this task is learnt using this subset of compact weights and the free weights. The network is grown during the training if the performance does not reaching a satisfactory level, providing more free weights. The training is followed by pruning to compact the newly learned weights. This method is useful in scenarios with less number but larger tasks. 

\subsubsection{Similarity with other works}
Here we specifically mention some of the similar works in the literature and point out the differences with our ours. Our main difference is that we train a set of specialized classifiers dedicated to each class group while also making them able to detect if a sample is from previous classes.

Aljundi et al. \cite{aljundi2017expert} trained a specialized model for each new task and proposed a method for choosing the relevant one at inference time. Specifically, by training separate autoencoders describing each task's data distribution, during inference they are able to choose the most relevant model for which the corresponding autoencoder has the least reconstruction error. After an autoencoder for a task is trained, it is used to select the most relevant task to the current task. Then the classifier model is trained based on the most related tasks.

AR1\cite{maltoni2019AR1} trains a linear classifier on top of a feature extractor. In each phase, a new linear classifier is trained for the current group of classes. Then a mean-normalization is applied to the weights of this classifier to eliminate the prediction bias. In testing, the prediction of the model is the softmax over outputs of all classifiers. The difference with our method is that we use deeper classifiers, while they do not. On the other hand, we use memory buffer, but AR1 doesn't use one.

\section{Methods}
Our purpose is to sequentially train on a number of disjoint datasets containing different classes. Relying on a pretrained deep feature extractor, we only consider training small networks on top of it. The traditional way is to construct and train a single head with expanding output units to facilitate the prediction of new classes. However, despite the replay mechanism, the single head is still not sufficiently equipped against catastrophic forgetting. We improve the situation by instead training much smaller classifiers, one at a time that are specific only to new classes in the training session. For a descriptive diagram of our method, please see Fig. \ref{fig:diagram}. In the following paragraph, we further formalize the training setting and details of our method.

In class-incremental learning we aim to train a network on a sequence of datasets $\mathcal{D}_i = {(x_i^j, y_i^j)}_{j=1}^{n_i}$ with inputs $x_i^j$ and labels $y_j^i \in \mathcal{Y}_i$, where $\mathcal{Y}_i \cap \mathcal{Y}_k = \emptyset$ for any $i \neq k$. Note that in the training session $i$ only the dataset $\mathcal{D}_i$ is available. Suppose we are allowed to store a small number of examples from the current training session in a limited memory buffer. We denote the samples in the memory buffer by $\mathcal{M}_i$ and its capacity by $B$, thus $\mathcal{M}_i \leq B$. The memory can be updated with new examples at each training session. Having access to past examples through the memory buffer, the training data for the $i$-th session will then consist of $\mathcal{D}_i \cup \mathcal{M}_i$.

Let $\mathbf{F}(\cdot)$ be the feature extraction network. Our approach is that at $i$-th training session we create a new classifier $\mathbf{C}_i$ network with $|\mathcal{Y}_i|$ output units that classifies between the new classes. The final classification is done through the final layer $S(\cdot) = \text{Softmax}(\oplus(\cdot))$ which simply concatenates the outputs of all classifiers and then applies the softmax. The objective is to minimize the cross entropy loss between the outputs of the final layer and the true target over all training samples:
$$
\mathcal{L}_{\theta_i}(x, y) 
= - \sum_{j}^{C}t_j\log(S(
c_0, \ldots , c_i
)_j) \\
$$
where
\begin{align*}
c_k &= \mathbf{C}_k(\mathbf{F}(x)) \quad \text{for} \quad  k = 0, \ldots, j \\
\theta_i &- \text{ parameters of the last classifier, } \mathbf{C}_i
\\
t &- \text{the one-hot encoded vector of target $y$}
\\
C &= \sum_{k=1}^{i} |\mathcal{Y}_k|  \quad \text{(i.e. the number of classes seen so far)} \\
\end{align*}

Note that only the last classifier is trained, while the loss is a function of the outputs of all classifiers. In this way, the last classifier is adjusting itself to respect the prediction of previous classifiers. In other words, it is learning to produce higher output values for the samples belonging to its feature space (the new classes in the current training session) while suppressing itself for the samples belonging to previous classifiers. Although all of the classifiers are in a sense ``partial", i.e. they can only predict the classes belonging to them, they also see previous classes during the training as ``negative" examples acting like a regularizer which makes them even stronger and robust. We also emphasize that since the previous classifiers are frozen, it eliminates the problem of ``forgetting'' for them. 

\subsection{Memory buffer}
As the memory should be kept updated to include new samples, a sampling strategy has to decide which examples should be selected or removed. We adopt the greedy sampling approach described in \cite{prabhu2020greedy}. It randomly replaces some of the old examples in the memory with the new ones, while trying to satisfy the balancing constraint, that is, to maintain an equal number of examples for each class.

\subsection{Training} 
Training consists of multiple sessions. At the beginning of each training session, a classifier is constructed which has the output size equal to the number of classes in the training data. Training objective is to learn the parameters of this new classifier. We train it by mini-batch gradient descent where each mini-batch contains an equal number of samples from the current dataset and the memory buffer. At the end of each training session the memory buffer is updated with new samples.

\subsection{Early Stopping}
As the number of examples per class in the memory keeps decreasing over time, the model becomes prone to overfitting. Therefore, an early stopping mechanism is essential. We propose that a portion of the data should be held for validation. Apart from splitting the incoming data into train and validation parts, we also dedicate a part of the memory for validation samples. We make sure that the part of the memory for training gets updated only with the samples in the training set, and similarly the validation part of the memory gets updated only with the samples in the validation set. 

\subsection{Bias correction layer}
Since the mini-batches contain nonequal number of samples from the old and new classes, it poses a class imbalance problem. This problem has been first addressed in \cite{wu2019BiC} and the authors proposed a simple yet effective solution, called Bias Correction (BiC). It is a layer containing only two parameters and applies a linear transformation to the output logits belonging to new classes $\mathcal{Y}_i$ as follows:
\begin{align*}
q_{k}=\left\{\begin{array}{lr}
\alpha o_{k}+\beta & k \in \mathcal{Y}_i \\
o_{k} &  otherwise
 \end{array}\right.
\end{align*}
where $\alpha$ and $\beta$ are the bias parameters and $o_k$ is the output logits of the final layer.
The BiC layer is trained separately at the end of each training session with a small amount of data as it contains only two parameters.

\section{Experimental Results}
We consider the class-incremental learning scenario on CIFAR100 dataset by splitting it into 5, 10, and 20 disjoint parts, each part containing 20, 10, and 5 classes respectively. In all of our experiments, we use the same class ordering which is obtained by random shuffling. We also run experiments with different memory sizes. We compare our method against two state-of-the-art baselines: ER equipped with BiC, and GDumb. 

\subsection{Implementation details}
For feature extraction, we use a EfficientNet-B0 network\cite{tan2019efficientnet} that is pretrained on ImageNet. We remove from it the final convolution layer, the output layer and the final MBConv block. Since this feature extractor accepts inputs of size $224\times224$, we resize the CIFAR100 images which are $32\times32$ to match the input size. The feature extractor is kept frozen in all experiments of our method and the baselines. For classifiers, we use a single $1\times 1$ convolution layer followed by global average pooling, and a dense layer. In all experiments the classifiers have the same architecture except the number of filters in the convolution layer and the number of units in the final dense layer. Overall, we have made sure that our method does not use more trainable parameters in total than the baselines (See table \ref{tab:parameters}). 

In all experiments, we stop a training session when the validation loss does not improve for 10 consecutive epochs. We have dedicated 10\% of the data for validation. For all methods we start by learning rate of 0.01. For our method we decrease the learning rate when the validation loss does not improve for 3 consecutive epochs. For GDumb and ER we apply exponential decay to learning with rate 0.95. At the end of each training session, just before testing,  we train the BiC layer to remove the prediction bias towards later classes. Since the whole purpose of BiC is to remove prediction bias, we will train on it only on the validation part of the memory buffer(after it has been updated to include all seen classes) which the model itself has never trained on.

\begin{figure}
    \begin{center}
        \input{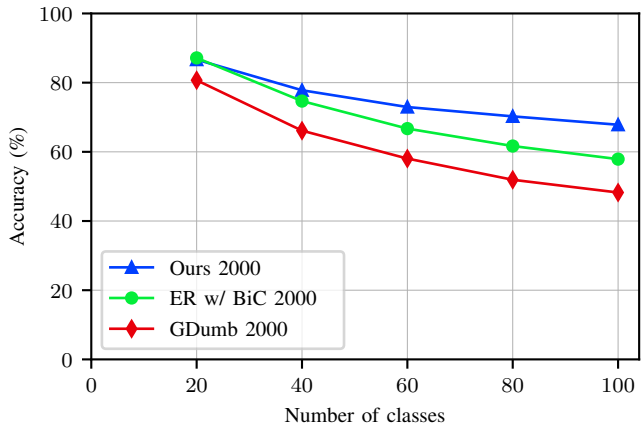}
    \end{center}
    
\caption{Accuracies at the end of each training session. CIFAR100 split into 5 parts.}
\label{fig:cifar100-5}
\end{figure}

\begin{figure}

    \begin{center}
        \input{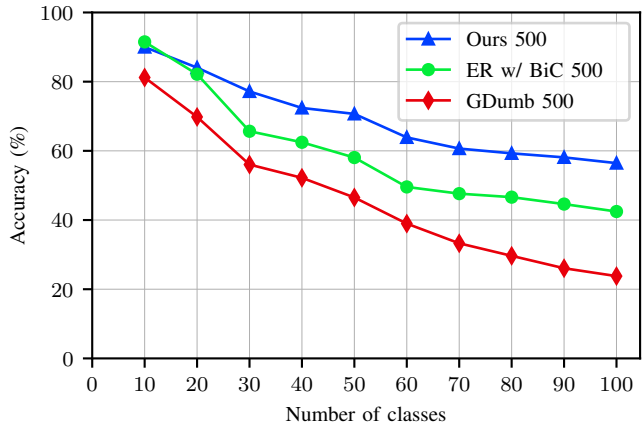}
    \end{center}
    
    \begin{center}
        \input{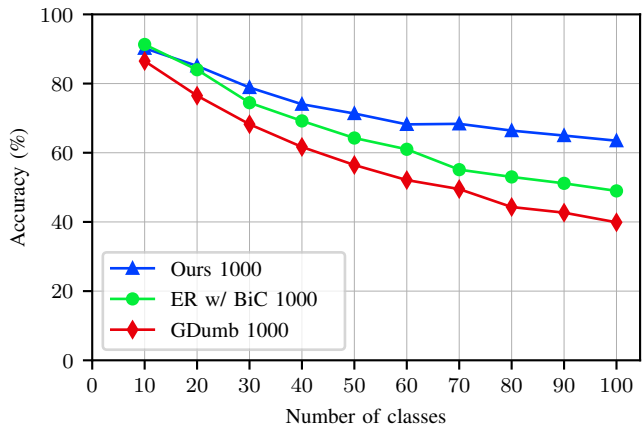}
    \end{center}
    
    \begin{center}
        \input{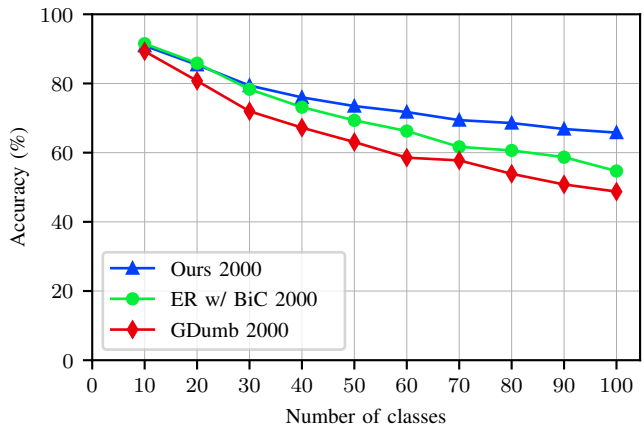}
    \end{center}

\caption{Accuracies at the end of each training session. CIFAR100 split into 10.}
\label{fig:cifar100-10}

\end{figure}

\begin{figure}
    \begin{center}
        \input{cifar100-20-1000.pgf}
    \end{center}
    
    \begin{center}
        \input{cifar100-20-2000.pgf}
    \end{center}
    
\caption{Accuracies at the end of each training session. CIFAR100 split into 20 parts.}
\label{fig:cifar100-20}

\end{figure}

\begin{table}[]
\begin{center}
    
\caption{Accuracies at the end of the training. CIFAR100 split into 5, 10, and 20 parts.}
\label{tab:cifar100}
\begin{tabular}{|l|l|lll|ll|}
\hline
Splits    & \multicolumn{1}{c|}{5}    & \multicolumn{3}{c|}{10}                                                        & \multicolumn{2}{c|}{20}                              \\ \hline
Memory    & \multicolumn{1}{c|}{2000} & \multicolumn{1}{c}{500} & \multicolumn{1}{c}{1000} & \multicolumn{1}{c|}{2000} & \multicolumn{1}{c}{1000} & \multicolumn{1}{c|}{2000} \\ \hline
GDumb     & 48.24                     & 23.79                   & 39.89                    & 48.74                     & 39.57                    & 47.83                     \\
ER w/ BiC & 57.89                     & 42.46                   & 48.97                    & 54.68                     & 48.78                    & 55.23                     \\
Ours      & \textbf{67.83}            & \textbf{56.45}          & \textbf{63.48}           & \textbf{65.79}            & \textbf{61.60}           & \textbf{63.25}            \\ \hline
\end{tabular}

\end{center}
\end{table}

\begin{table}[]
\begin{center}
    
\caption{Number of trainable parameters in each of the experiment settings, when training on CIFAR100 split into 5, 10, and 20 parts.}
\label{tab:parameters}
\begin{tabular}{|l|l|l|l|}
\hline
\multirow{2}{*}{Method} & \multicolumn{3}{c|}{\# trainable parameters} \\ \cline{2-4} 
                        & 5 splits       & 10 splits    & 20 splits    \\ \hline
GDumb                   & 118K           & 118K         & 118K         \\
ER                      & 118K           & 118K         & 118K         \\
Ours                    & 17K $\to$ 86K  & 8K $\to$ 82K & 4K $\to$ 80K \\ \hline
\end{tabular}
\end{center}
\end{table}

\subsection{Baselines}
GDumb and ER use the same architecture, they only differ in training. Gdumb only trains on memory samples. It updates the memory buffer at the beginning of each training session and then trains only on the memory buffer discarding the rest of the data. We also train a BiC layer at the end of each training session of ER. BiC layer is not necessary for GDumb due to the fact that it trains on balanced dataset, i.e., the memory buffer.

\subsection{Analysis}
We test the models at the end of each training session over all classes that have been encountered so far. In table \ref{tab:cifar100} we report the  accuracy at the end of the training in three different settings with varying memory buffer sizes (500, 1000, and 2000) and dataset splitting into 5 (Fig. \ref{fig:cifar100-5}), 10 (Fig. \ref{fig:cifar100-10}), and 20 (Fig. \ref{fig:cifar100-20}) parts. In all settings we see a large gap in accuracy between our method and the second best method, ER. ER always outperforms GDumbs, as expected, because it trains on all available data and the early stopping mechanism that we introduced here prevents it from unnecessary training which causes more forgetting of past data. 

We observe that the gap in accuracy tends to get larger when we shrink the memory buffer size. Our method still reaches a remarkable performance of 56.45\% when memory size is 500 leaving a gap of 14\% relative to the next method, ER. On the other hand, we see a large performance drop of GDumb. This is mainly because it is highly dependent on the memory size as it trains only on the memory samples. However, GDumb shows the least difference in performance when training with different dataset splittings, meaning that it might have an advantage in settings more close to online learning.

\subsection{Ablation studies}
We conduct experiments to see if all components of our method are indeed important. The first component is that we freeze the classifiers after we train them so that in this way we believe the problem of ``forgetting" is eliminated. Therefore we have conducted an experiment where we let our method to keep updating all classifiers. The second component is the additional BiC layer. Looking at the confusion matrices, we have observed a prediction bias towards the last group of classes which was a signal to incorporate BiC layer. We also present the results here where our method does not use a BiC layer. These experiments are conducted in the case of CIFAR100 split into 10 classes, and a memory buffer size of 2000. The reported results are in Table \ref{tab:cifar100-ablation}. We can see the benefit of using BiC layer, which confirms that our method would have some prediction bias without it. On the other hand, we can see a large drop in accuracy  (65.79\% $\to$ 57.04\%) when we allow updating our classifiers. Nevertheless, still the performance stays above the other two methods (looking at Table \ref{tab:cifar100}, 54.68\% and 48.74\% of ER and GDumb, respectively).

\begin{table}[]
\caption{Ablation studies testing our method in two alternative forms: (1) without freezing the classifiers, and (2) without using BiC layer.}

\begin{center}
    
\label{tab:cifar100-ablation}
\begin{tabular}{|l|l|}
\hline
Method              & \multicolumn{1}{c|}{Accuracy \%} \\ \hline
Ours w/out freezing & 57.04                            \\ \hline
Ours w/out BiC      & 62.86                            \\ \hline
Ours                & 65.79                            \\ \hline
\end{tabular}

\end{center}

\end{table}

\section{Conclusion}
In this paper, we proposed a new approach for class-incremental learning. We tackled the problem by training a separate classifier for each new group of classes. By freezing these classifiers after they have been trained, we have limited the problem of ``forgetting" and achieved big improvements over strong SOTA baselines. Our method consistently achieved better performance when tested on CIFAR100 learning different number of classes at a time. 

\bibliographystyle{IEEEtran}
\bibliography{main.bib}

\end{document}